\documentclass[conference]{IEEEtran}
\IEEEoverridecommandlockouts
\usepackage{cite}
\usepackage{amsmath,amssymb,amsfonts}
\usepackage{algorithmic}
\usepackage{graphicx}
\usepackage{textcomp}
\usepackage{xcolor}
\usepackage{multirow}
\usepackage{placeins}
\usepackage[bookmarks=false]{hyperref}
\newcommand{\link}[1]{{\color{blue}\href{#1}{#1}}}

\def\BibTeX{{\rm B\kern-.05em{\sc i\kern-.025em b}\kern-.08em
    T\kern-.1667em\lower.7ex\hbox{E}\kern-.125emX}}
\begin{document}

\title{Large-Scale Intelligent Microservices
\thanks{Thanks to Microsoft and MIT Systems that Learn Consortium for their support on this project}
}

\author{\IEEEauthorblockN{Mark Hamilton}
\IEEEauthorblockA{\textit{Cognitive Services Research} \\
\textit{Microsoft, MIT}\\
Boston, USA \\
marhamil@microsoft.com}
\and
\IEEEauthorblockN{Nick Gonsalves}
\IEEEauthorblockA{\textit{Cognitive Services Research} \\
\textit{Microsoft}\\
Boston, USA\\
nigons@microsoft.com}
\and
\IEEEauthorblockN{Christina Lee}
\IEEEauthorblockA{\textit{Applied AI} \\
\textit{Microsoft}\\
Bellevue, USA\\
chril@microsoft.com}
\and
\IEEEauthorblockN{Anand Raman}
\IEEEauthorblockA{\textit{Applied AI} \\
\textit{Microsoft}\\
Bellevue, USA\\
aram@microsoft.com}
\and
\IEEEauthorblockN{Brendan Walsh}
\IEEEauthorblockA{\textit{Applied AI} \\
\textit{Microsoft}\\
Bellevue, USA\\
brwals@microsoft.com}
\and
\IEEEauthorblockN{Siddhartha Prasad}
\IEEEauthorblockA{\textit{Applied AI} \\
\textit{Microsoft}\\
Bellevue, USA\\
siprasa@microsoft.com}
\and
\IEEEauthorblockN{Dalitso Banda}
\IEEEauthorblockA{\textit{Grey Systems Lab} \\
\textit{Microsoft}\\
Boston, USA\\
dbanda@microsoft.com}
\and
\IEEEauthorblockN{Lucy Zhang}
\IEEEauthorblockA{\textit{Applied AI} \\
\textit{Microsoft, MIT}\\
Boston, USA\\
zhangly@mit.edu}
\and
\IEEEauthorblockN{Mei Gao}
\IEEEauthorblockA{\textit{Cognitive Services Research} \\
\textit{Microsoft}\\
Boston, USA\\
xuemei.gao@microsoft.com}
\and
\IEEEauthorblockN{Lei Zhang}
\IEEEauthorblockA{\textit{Cognitive Services Research} \\
\textit{Microsoft}\\
Bellevue, USA\\
leizhang@microsoft.com}
\and
\IEEEauthorblockN{\centerline{William T. Freeman}
\IEEEauthorblockA{\textit{CSAIL} \\
\textit{MIT}\\
Boston, USA\\
billf@mit.edu}}
}

\maketitle

\begin{abstract}
Deploying Machine Learning (ML) algorithms within databases is a challenge due to the varied computational footprints of modern ML algorithms and the myriad of database technologies each with their own restrictive syntax. We introduce an Apache Spark-based micro-service orchestration framework that extends database operations to include web service primitives. Our system can orchestrate web services across hundreds of machines and takes full advantage of cluster, thread, and asynchronous parallelism. Using this framework, we provide large scale clients for intelligent services such as speech, vision, search, anomaly detection, and text analysis.  This allows users to integrate ready-to-use intelligence into any datastore with an Apache Spark connector. To eliminate the majority of overhead from network communication, we also introduce a low-latency containerized version of our architecture. Finally, we demonstrate that the services we investigate are competitive on a variety of benchmarks, and present two applications of this framework to create intelligent search engines, and real time auto race analytics systems.

\end{abstract}

\begin{IEEEkeywords}
services, spark, map-reduce, SQL, streaming, cognitive services, text analytics, vision, speech, anomaly detection, containers, micro-services.
\end{IEEEkeywords}

\section{Introduction}

\begin{figure*}[ht]
\vskip 0.1in
\begin{center}
\centerline{
\includegraphics[width=.90\linewidth]{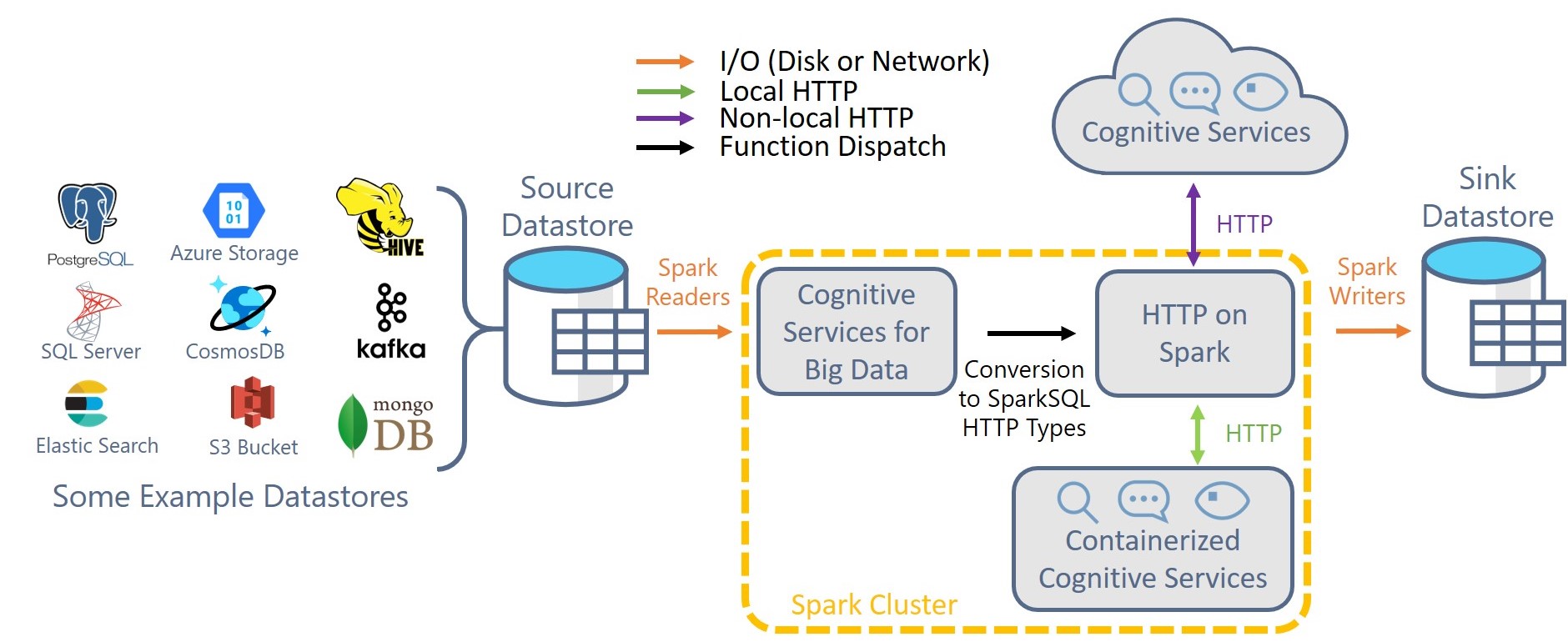}
}
\caption{End to end architecture diagram for the Cognitive Services for Big Data. Using this architecture application developers can enrich their databases with both containerized and web-based intelligent services.}
\label{fig:overview}
\end{center}
\end{figure*}
 
Databases are a backbone of modern computing infrastructure. Though most of the world's data is housed in databases, domain-specific programming models and limited APIs can make it difficult to perform machine learning (ML) within these systems. The growing complexity of modern ML frameworks only exacerbates the problem. Today's ML frameworks are resource-intensive and can require specialized hardware such as Graphical Processing Units (GPUs), Tensor Processing Units (TPUs) \cite{jouppi2017datacenter}, and Field Programmable Gate Arrays (FPGAs) \cite{ebeling1993field}. Often, this hardware is expensive, and steep costs dictate that applications and users share resources efficiently. This leads many system designers to use multi-tenant services to abstract, share, and independently scale ML components. This work investigates the extent to which these two paradigms; databases and intelligent services, can be efficiently and \textit{simply} integrated. Our primary aim is to present a new architecture and demonstrate its efficiency. We also aim to show that the algorithms we consider are competitive within the broader landscape of managed intelligent services. In summary, this work contributes:

\begin{itemize}
    \item A distributed and database-centric micro-service orchestration framework built on Apache Spark.
    \item Large-scale intelligent service clients for enriching a broad class of databases with managed intelligent algorithms.
    \item Low-connectivity and low-latency variants of these architectures using containerized intelligent services.
    \item The first systematic comparison of the quality of intelligent cloud services across major cloud providers.
\end{itemize}

\section{Background}

The field has seen a proliferation of technologies that fill ``database-like'' roles in the computing ecosystem. These technologies span a wide variety of APIs and types of queries. Some important families are SQL-based (SQL Server \cite{sql-server}, BigQuery \cite{sato2012inside}, PostgreSQL \cite{douglas2003postgresql}), NoSQL-based (MongoDB \cite{chodorow2013mongodb}, CosmosDB \cite{cosmosdb}, Neo4j \cite{neo4j}), storage-based (Azure Storage \cite{calder2011windows}, AWS S3 \cite{s3}, Google LFS \cite{ghemawat2003google}), Search-based (ElasticSearch \cite{gormley2015elasticsearch}, Azure Cognitive Search), and stream-based (Kafka \cite{kreps2011kafka}, EventHub, RabbitMQ \cite{richardson2008introduction}). Due to the large diversity of frameworks, it's natural to seek a single platform with modular connectors to each other system. One of the most popular systems to abstract operations over these databases is Apache Spark, which provides a distributed and fault-tolerant data-store application programming interface (API) over these and other systems. Apache Spark combines the functionality of SQL and Map-Reduce \cite{mapreduce} and operates on large elastic pools of machines. Since its introduction, Apache Spark has added multi-lingual bindings in several popular programming languages, stream processing \cite{armbrust2018structured}, graph-processing \cite{xin2013graphx, dave2016graphframes}, and machine learning \cite{sparkml}. This broad framework support and API flexibility enable us to focus on integrating intelligent services into Apache Spark, as any integration will inherit Apache Spark's broad database interoperability. 

With Apache Spark serving as a uniform abstraction over datastores we can turn our attention to abstracting over machine learning systems. There are a wide variety of frameworks, model formats, and programming languages a scientist can choose from when building ML algorithms \cite{tensorflow, pytorch, sklearn, malohlava2016machine, carpenter2017stan, salvatier2016probabilistic, witten1999weka, ke2017lightgbm}. This plurality of frameworks and the high cost of hardware acceleration lead many to use service-based architectures. Service-based architectures compartmentalize logical units of computation into ``services'' that can communicate with each other. Often, multiple applications can use the same back-end service, allowing system designers to minimize machine downtime and maximize efficiency. Furthermore, abstracting logic into services allows systems to leverage multiple frameworks, languages, or machines. There are several different protocols that modern services use to pass information, such as RPC \cite{srinivasan1995rpc}, gRPC \cite{grpc}, WebRTC \cite{rescorla2013webrtc}, QUIC \cite{langley2017quic}, WebSocket \cite{fette2011websocket}, and IPFS \cite{benet2014ipfs}. However, the most widely used is the Hypertext Transfer Protocol (HTTP) \cite{fielding1999hypertext}. This protocol forms the basis of websites, the Open API specification \cite{openapi2017openapi}, and is supported by popular service frameworks like Flask \cite{grinberg2018flask}, nginx \cite{nedelcu2010nginx}, Jetty \cite{jetty}, Node.js \cite{tilkov2010node}, and Apache Webserver \cite{hu1999measurement}, and ASP.NET \cite{guthrie2007asp}. Though newer protocols like gRPC are better optimized for streaming large quantities of data, HTTP still dominates the broader community. In particular, Google Cloud, Microsoft Azure, Amazon Web Services (AWS), IBM Watson, and others use HTTP to make their intelligent web services available to users. 

\begin{figure*}[ht]
\vskip 0.1in
\begin{center}
\centerline{
\includegraphics[width=.90\linewidth]{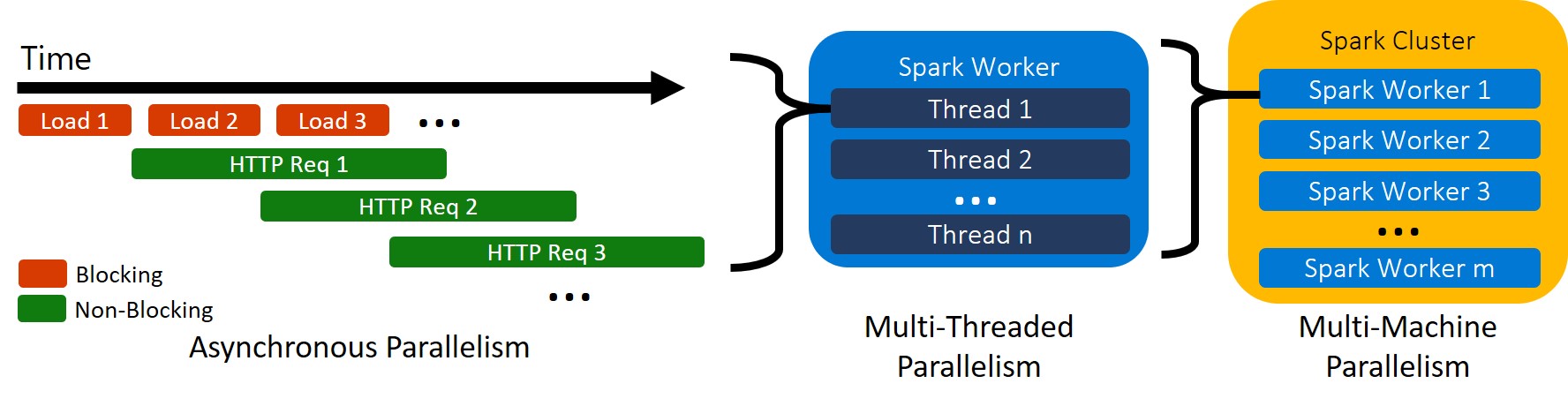}
}
\caption{Diagram visualizing the different forms of parallelism in HTTP on Spark. We introduce asynchronous parallelism to Apache Spark to dramatically improve throughput without affecting Apache Spark's ambient thread and machine parallelism.}
\label{fig:parallelism}
\end{center}
\end{figure*}

\section{Related Work}

There is a large body of work on creating high-performance web clients for a variety of applications. Swagger Codegen \cite{swagger-codegen} uses the Open API specification to automatically generate idiomatic HTTP clients in a wide array of target languages. Swagger Codegen focuses on making a single web service call and does not focus on how to integrate the clients with databases which separates it from this work. The GRequests \cite{grequests} project aims to make it easy to use Python's GEvent async framework to send web requests using asynchronous parallelism.  This framework makes it easy to transform Python ``Requests'' \cite{reitz2020requests} code into a more performant form, but does not tackle the issue of integrating with Python's DataFrame API, Pandas \cite{mckinney2011pandas}. In a similar vein, the Twisted project \cite{fettig2005twisted} aims to provide an asynchronous and event-driven API to Python for web serving and web clients. Again, this work does not explicitly tackle the problem of multi-node database integration. In the JVM ecosystem, Akka \cite{vernon2015reactive} has become a popular framework for actor-based and event-driven asynchronous parallelism and streaming. Like Twisted and Grequests, this framework does not focus on integration with databases, but does however scale to multiple machines using the ``akka-cluster'' package. Tools like Locust \cite{locust} allow for high throughput clients but focus on load testing as opposed to ETL and databases. Another vein of related work comes from the workflow and process automation community. Tools like Logic Apps \cite{logic-apps} and App Connect \cite{app-connect} allow users to pipe data through a graph of computations including intelligent services. These tools achieve similar connectivity to our work, but they do not standardize on the DataFrame API and they do not reap the benefits of the Spark's Catalyst optimizer such as query re-ordering, or query push-down which can speed code by orders of magnitude by reducing unnecessary IO. Furthermore, many process automation providers expose their systems as Graphical User Interfaces or as black-box services, which limits extensibility and prohibits native integration with other software.

In addition to work in high-performance clients, there has been several standards proposed for unified protocols for database communication. One of the most successful is the Open Database Connectivity (ODBC) standard \cite{signore1995odbc} which allows for applications to access a wide variety of datastores using the same API. Apache Spark supports ODBC connections a data sources and sinks which allows our work to extend to databases that adhere to this protocol. 

Finally, there is a large body of work on machine learning operationalization frameworks. Two that are particularly relevant are Tensorflow Serving \cite{olston2017tensorflow} and Kubeflow Serving \cite{kubeflow2020}. These tools allow for the fast deployment of a variety of models as gRPC endpoints and are particularly well integrated with the Kubernetes stack. gRPC endpoints are particularly well suited for high-throughput usage due to their bidirectional communication protocol and compression scheme for sending repeated records. TFServing has tools to generate clients automatically from inspecting through Protocol Buffers \cite{varda2008protocol}. These frameworks enable integration of machine learning components efficiently into other systems, but only if they leverage gRPC protocol, which major intelligent service providers do not yet widely adopt. Nevertheless, we hope to explore other protocols in addition to HTTP in future work.

\section{Architecture}

\begin{figure*}[ht]
\vskip 0.1in
\begin{center}
\centerline{
\includegraphics[width=.80\linewidth]{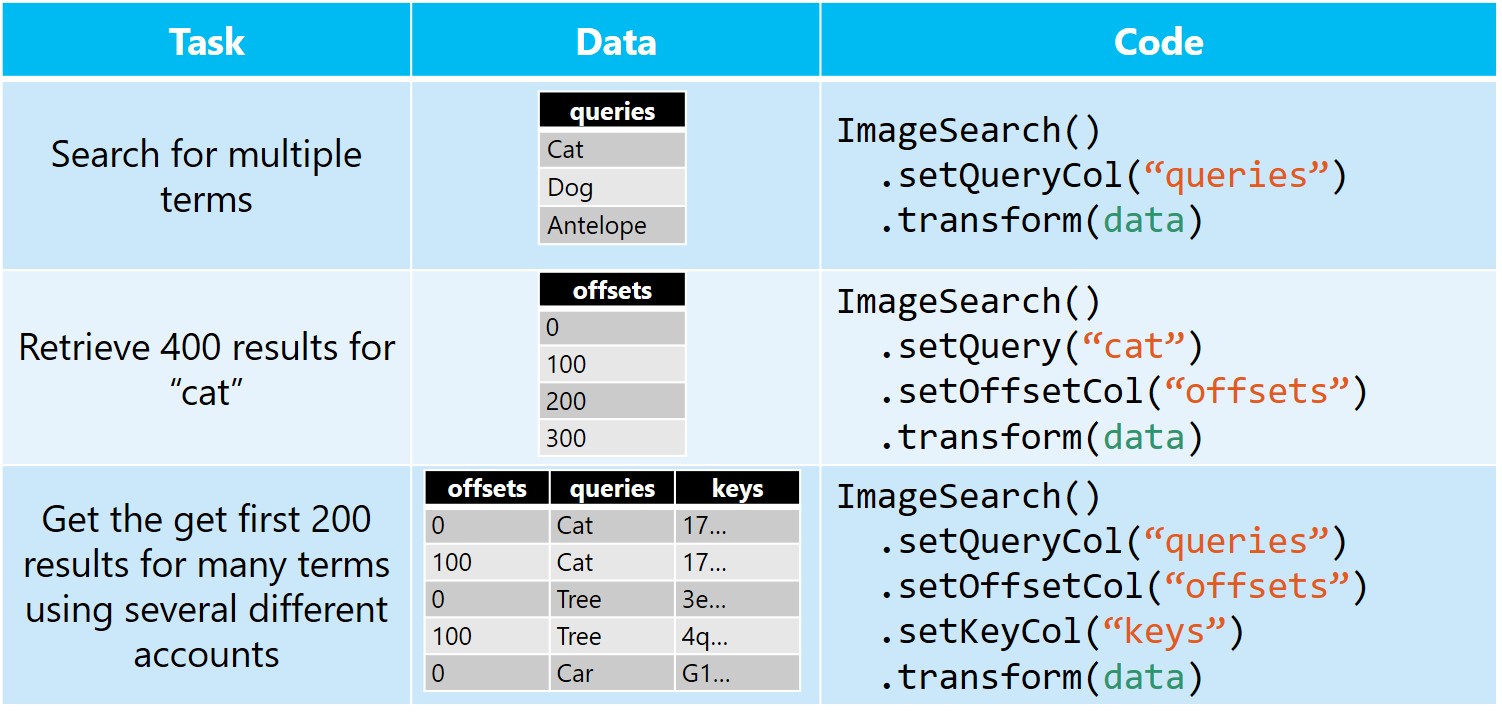}
}
\caption{Example usage of the Cognitive Services for Big Data. Our fluent API allows a single client to apply to a variety of different queries without bloated initialization code or a large DataFrame of unnecessary parameters. }
\label{fig:api-example}
\end{center}
\end{figure*}

\subsection{HTTP on Spark}

The first layer of abstraction in our architecture is the networking layer we call ``HTTP on Spark''. This layer integrates the full HTTP communication protocol with Spark SQL. More specifically, we use Spark SQL's type constructors to build new SQL types for HTTP requests and responses. By encoding the protocol in the SQL type system directly, the full breadth of queries and optimization strategies apply to the new HTTP types. In particular, filters, order-bys, logical predicates, shuffles, and other operations can be moved through this type transparently with the Catalyst Optimizer \cite{armbrust2015spark}. Representing these objects as SQL types also allows any of Spark's language bindings such as PySpark, SparklyR, and Spark.NET to work with the new types. With a consistent representation of the HTTP protocol, we add Spark primitives for efficient HTTP communication. At large scales, HTTP communication becomes more complex. Sending many requests to a single back-end increases the probability of malformed requests, transient errors, rate limiting, and server-side issues. We ensure that HTTP on Spark does not overwhelm rate-limited services by leveraging HTTP ``Retry-After'' headers and rate-limit status codes as signals for back-pressure. Furthermore, we guard against flaky connections and services with several layers of exponential-back-off in both the HTTP layer and the application layer. 

HTTP on Spark allows Spark users to leverage the parallel networking capabilities of their cluster to integrate any local, Docker, or web service with a variety of data sources and sinks. Our integration of the HTTP protocol creates a simple and principled way to integrate arbitrary computation frameworks into the Spark ecosystem that is independent from their computational requirements. When combined with Spark's ML Pipelines, users can chain services together, allowing Spark to function as a distributed micro-service orchestrator.

\subsection{Asynchronous Parallelism in Spark}

Though retries and back-pressure can help maximize throughput and balance network traffic across a pipeline, there is still a considerable amount of inefficiency with synchronous HTTP communication. More specifically, synchronous communication under-utilizes network IO parallelism and wastes time during server-side computation. However, Apache Spark's form of parallelism does not offer abstractions to overlap the scheduling of tasks and computation. To address these issues, we introduce a new method for adding asynchronous parallelism into Apache Spark. More specifically, we provide an additional DataFrame operation that behaves like ``mapPartitions'' but provides a buffered iterator of row futures instead of a standard iterator of rows. With this, each Spark worker thread can multi-task during non-blocking operations such as waiting for an HTTP response. Furthermore, this contribution does not interfere with Spark's organization of threads and processes and generalizes naturally to batch, streaming, and real-time computation scenarios. We illustrate the different forms of parallelism in Figure \ref{fig:parallelism}. Our addition of tunable asynchronous parallelism can increase web client throughput by orders of magnitude as demonstrated by Figure \ref{fig:async-parallelism}.

\subsection{The Cognitive Services for Big Data}

With a performant HTTP networking layer in place, we can turn our attention to providing and integrating a wide variety of intelligent services to provide easy and performant ML for a broad class of use-cases. More specifically, we provide a wide variety of guaranteed-up-time multi-tenant intelligent services called the ``Azure Cognitive Services''. These services span a wide array of algorithms including text sentiment classification, personally identifying information removal, named entity detection, time-series anomaly detection, visual object recognition, visual captioning and tagging, face recognition and detection, face verification, content moderation, multi-lingual speech-to-text, web-based image search, and several others. We guarantee up-time with a variety of service-level-agreements (SLAs) and maintain dedicated deployments in over 56 globally distributed regions to minimize request latency independent of geographic location.

Using HTTP on Spark as a shared layer of abstraction, we provide distributed and high-throughput clients for each Cognitive Service to enable bringing intelligent services to big-data systems and databases. Our contributions adhere to the SparkML API for modular-reuse, compositionality, and broader ecosystem compatibility. We refer to these as the ``Cognitive Services for Big Data''. These clients benefit from the variety of networking optimizations in HTTP on Spark and contain additional optimizations for services that naturally support batching. The Cognitive Services for Big Data employ these optimizations under the hood and abstract away low-level networking details to allow database and distributed application developers to focus on their applications instead of the challenges associated with of robust service communication. Additionally, we continually improve the cloud-hosted Cognitive Services with zero-downtime deployments so database intelligence will continuously improve \textit{without} re-deploying pipelines or application code.

\begin{figure*}[ht]
\vskip 0.1in
\begin{center}
\centerline{
\includegraphics[width=.80\linewidth]{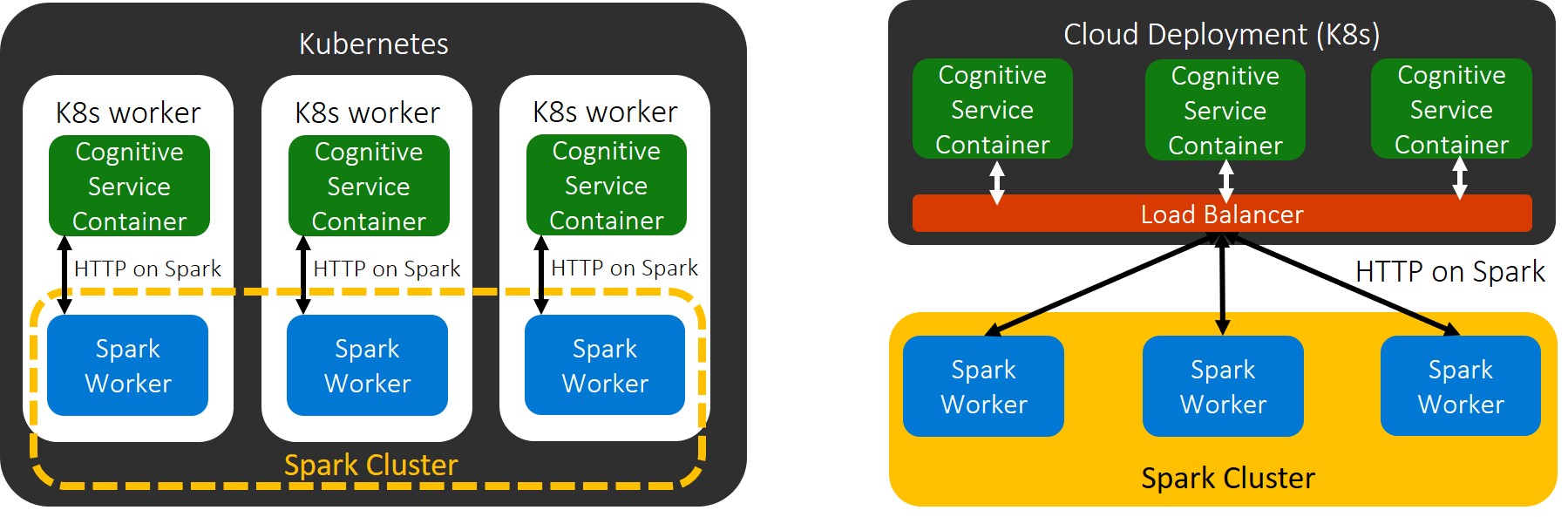}
}
\caption{Architecture diagram of our integration of cloud and containerized Cognitive Services. For cloud services, we actively maintain multi-tenant Kubernetes clusters so that users do not need to manage this infrastructure. Architecture depicted on Kubernetes, but any container orchestrator could deploy the same. Note that we omit Kubernetes and Spark head nodes for simplicity.}
\label{fig:k8s-architecture}
\end{center}
\end{figure*}

\subsection{Fluent Design and API}

 The Cognitive Services for Big Data transform distributed tables of data, called Spark ``DataFrames'' \cite{sparksql}, by adding new columns containing the results of machine learning transformations. In Apache Spark, computations on DataFrames are rigorously optimized and ``schema-checked'' to improve performance and usability. Describing computations with DataFrames simplifies programming with heterogeneous structured data sources and allows users to build complex workflows with many fewer lines of code than ``loop-centric'' programming.  As a result, many programming languages feature a DataFrame API \cite{mckinney2011pandas,R,wickham2015dplyr}, and there has been a proliferation of DataFrame-based APIs for machine learning \cite{sklearn-pandas,sparkml,hunter2016tensorframes,gulli2017deep}, visualization \cite{waskom2015seaborn,plotly}, and other workloads. In this work we provide a DataFrame-based API for networking operations. 
 
Parametrizing complex and arbitrary machine learning tasks can be a challenge when building a DataFrame API. More specifically, each Cognitive Service can take in a variety of different parameters. For example, a sentiment analysis algorithm might require text to analyze, a language, and a service key. It is not necessarily clear how to differentiate between ``data-plane'' parameters, like the text, which are parameterized by entire DataFrame columns and ``control-plane'' parameters such as a service key, which are typically shared across every service call. If we require that users to supply all algorithm parameters through the ``data-plane'' the resulting system is maximally flexible. However, this requires users to make many temporary columns for their computations, which is both cumbersome and introduces unnecessary memory overhead. On the other hand, if we only treat a few parameters as input data, the API is simpler for a user, but may not be suitable for complex workloads. To handle a variety of use cases elegantly and efficiently we add new SparkML parameter types to allow a user to set query parameters with either an entire DataFrame column or a single value. Using this new parameter type, users can fill any argument of the web request from a DataFrame column for maximum flexibility, or from a static value for simplicity. 
 
 In scenarios where object initialization is complex, object constructors can become bloated and error prone. We leverage fluent design \cite{fowler2010domain} and object builder syntax so that the complexity of the initialization scales with the complexity of the query. When combined with code completion from a language server protocol \cite{lsp}, this allows users to quickly view all parameters and fill necessary parameters with type safety. We show several example uses and their corresponding code in Figure \ref{fig:api-example}. Finally, we note that we expose the Cognitive Services for Big Data and HTTP on Spark to PySpark, SparklyR \cite{venkataraman2016sparkr}, Scala, and Java using an automatic wrapper generation system to eliminate repeated, and tedious, code \cite{mmlspark}.

\subsection{Containerized Cognitive Services}
\label{sec:containers}

Web services are an important part of many software architectures because they allow multi-tenant resource sharing across any device that has an internet connection. However, due to networking and security constraints, many applications cannot maintain internet connectivity. Furthermore, large datasets can cause issues in bandwidth-constrained workflows, especially if applications require low latency. To avoid these pitfalls, we can instead deploy services near, or \textit{on} Spark workers. In the latter case, data never leaves executor machines, and local process communication is orders of magnitude faster than network communication as shown in Figure \ref{fig:containerized-latency}.

To bring the Cognitive Services to low and no-connectivity workloads and capitalize on the efficiency of local services, we provide Cognitive Services as Docker containers \cite{merkel2014docker}. These containers contain all necessary dependencies and can run on any system that runs Docker. This gives system designers control over service scale and network topology. To date, we are the only large-scale intelligent service provider to offer containerized services. Furthermore, the containers we release are the same containers we deploy as production cloud services, so their quality and APIs are identical. 
With containerized services, we can deploy intelligent algorithms next to data and reap the performance benefits of local networking while maintaining independence from the framework used by the intelligent algorithm. This approach allows us to keep the abstraction benefits of services without the overhead of network communication. Because containerized services have the same API as cloud services, we can leverage the same Cognitive Service clients for both web, local, and containerized scenarios just by changing the request URL. To simplify deployment we have contributed a helm chart  \cite{helm} for deploying a Spark-based microservice architecture with containerized Cognitive Services with a single command. Figure \ref{fig:k8s-architecture} illustrates both this containerized architecture and our cloud-service based architecture.

\section{Experiments}

To understand the performance characteristics of our system and how it relates to other approaches, we perform a variety of experiments using several text and vision services. We explore the design choices of our architecture and its speed relative to other approaches. To demonstrate that the algorithms under consideration are relevant and competitive we also perform an evaluation of semantic quality of the underlying intelligent services. To our knowledge, this fair analysis of service quality across the largest cloud service providers is the first of its kind published.

\begin{figure*}
    \centering
    \begin{minipage}{0.45\textwidth}
        \centering
        \includegraphics[width=\textwidth]{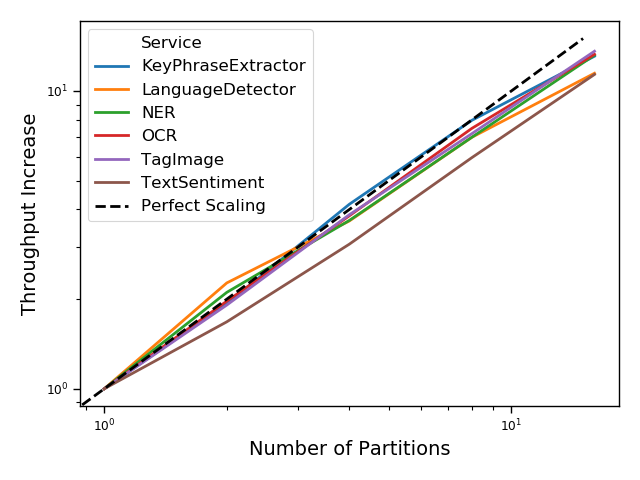} 
        \caption{Increase in request throughput as a function of Spark's inbuilt parallelism factor: Partitions.  Throughput increase represents the ratio of parallel throughput to single partition throughput. Performance scales linearly with Spark Workers so long as the back-end service can handle the load.}
        \label{fig:partition-parallelism}
    \end{minipage}
    \hfill
    \begin{minipage}{0.45\textwidth}
        \centering
        \includegraphics[width=\textwidth]{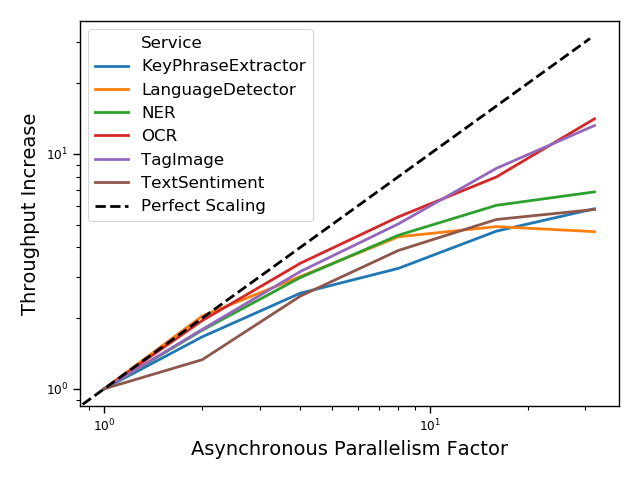} 
         \caption{Increase in request throughput as a function of Asynchronous Parallelism. We vary parallelism from 1 request per thread to a maximum of 32 requests per thread. Throughput increase represents the ratio of asynchronous parallel throughput to synchronous throughput. Asynchronous parallelism can increase throughput by an order of magnitude, especially for slower services like OCR and TagImages. However, these gains are not limitless as threads will eventually reach full working capacity.}
         \label{fig:async-parallelism}
    \end{minipage}
\end{figure*}

\subsection{Comparison to other Approaches}

To understand how our approach compares to other related tools, we measure throughput per compute node across several text and vision cloud services. This allows us to compare across both single and multi-node approaches in the community. In table \ref{table:competitors} we show the results of a throughput analysis across several high-throughput HTTP strategies. Our baselines include Python's ``requests'' library \cite{reitz2020requests} in a ``for'' loop (Requests), and within a Apache Spark User Defined Function (Requests + UDF). We also explore an asynchronous implementation of Python requests, Grequests, and the actor-based streaming and HTTP framework, ``Akka''. Though none of these approaches aim to solve the problem of integrating web services and databases explicitly, they are reasonable choices to consider for large scale HTTP client workloads. We compare these to the Cognitive Services for Big Data with (Ours + Async) and without (Ours) an asynchronous parallelism factor of 8 (up to 8 concurrent requests per thread). We find that both our baseline method and our asynchronous method improve throughput per compute node significantly compared to other approaches.

We measure throughput on a text dataset of 10k random sentences from the Project Gutenberg corpus \cite{gerlach2020standardized} and a vision dataset of 1000 random images from The Metropolitan Museum of Art Open Access collection \cite{Met}. Across these experiments we batch text with a batch size of 10, which is an upper limit for some services. In these and other throughput experiments, the content of the datasets is not the focus, as results depend primarily on the size of text and images and networking speeds. We chose these collections as they were representative of standard text and vision tasks, large enough to get reasonable steady-state throughput measurements, and helped us ensure our numbers were comparable across experiments. We perform this comparison on an Azure Databricks cluster (Spark 2.4.5, Scala 2.11, Python 3.6) with Standard\_D16s\_v3 head and worker nodes (64Gb RAM, 16 cores, 8000Mbps Network Bandwidth) and non-rate-limited services. Services tested include \textbf{Sentiment} Analysis, \textbf{N}amed \textbf{E}ntity \textbf{R}ecognition, \textbf{L}anguage \textbf{D}etection, \textbf{K}ey \textbf{P}hrase Extraction, \textbf{O}ptical \textbf{C}haracter \textbf{R}ecognition, and Image Tagging.  Numbers displayed represent throughput measured in rows per second per compute node to facilitate fair comparison. Using our approach can yield dramatic increases in throughput per compute node. We also note that requests and Grequests methods cannot scale to multiple workers without the use of another framework. Akka comparisons used an asynchronous parallelism factor of 32 to make. 

These results demonstrate that there is a significant benefit to using our approach due to its combination multi-core, multi-machine, and asynchronous parallelism. Furthermore, approaches leveraging compiled Scala code generally outperform the equivalent Python implementations using ``requests''. We also note that we have added a direct integration between HTTP on Spark and Python requests so that users with existing requests code-bases can leverage this acceleration and asynchronous parallelism with minimal code change.

\setlength{\tabcolsep}{4pt}

\begin{table}[ht]
\centering
\caption{Comparison of throughput (Rows per second per compute node) for several methods}
\begin{tabular}{|c|cccccc|}
\hline
\multirow{2}{*}{Method} & \multicolumn{6}{c|}{Service}                                                  \\ \cline{2-7} 
                        & Sentiment & NER & LD & KP & OCR & TagImage \\ \hline
Requests                & 3.0       & 3.1 & 3.3                & 3.2                   & 1.2 & 1.2      \\
Requests + UDF          & 37        & 40  & 48                 & 43                    & 19  & 19       \\
Grequests               & 4.1       & 4.1 & 4.2                & 4.0                   & 1.3 & 1.4      \\
Akka                    & 42        & 42  & 48                 & 36                    & 6   & 6        \\
Ours                    & 111       & 77  & 111                & 100                   & 25  & 24       \\
Ours + Async            & \textbf{200}       & \textbf{125} & \textbf{143}                & \textbf{167}                   & \textbf{125} & \textbf{100}      \\ \hline
\end{tabular}
\label{table:competitors}
\end{table}
\setlength{\tabcolsep}{6pt}

\subsection{Analysis of Parallelism}

To help us understand the performance of our architecture, we examine the scaling behavior with respect to both Spark's worker and thread parallelism (Number of Partitions) and our contribution of asynchronous parallelism. We measure throughput (rows per second) on the same two test datasets used in Table \ref{table:competitors}  to quantify the scaling behavior of our system as a function Spark's in-built parallelism factor (Partitions). We scale performance relative to the performance on a single spark partition to allow for comparisons across services. Figure \ref{fig:partition-parallelism} shows close to perfect scaling as we increase the number of Spark partitions. We expect this behavior because the task is naively parallelizable, and this trend will continue if the back-end services can handle the load.

We also perform the same analysis on our added asynchronous parallelism factor which allows each Spark thread to send multiple requests at once. In Figure \ref{fig:async-parallelism} we see that we can improve throughput by over an order of magnitude without introducing extra threads or hardware. However, this form of parallelism has its limits, as a single thread must work continuously to handle multiple requests and eventually saturates. Figure \ref{fig:async-parallelism} also shows that services with longer server-side computation times, such as image tagging and OCR, allow for more acceleration from asynchronous parallelism.

\begin{figure*}[ht]
\vskip 0.1in
\begin{center}
\centerline{
\includegraphics[width=.80\linewidth]{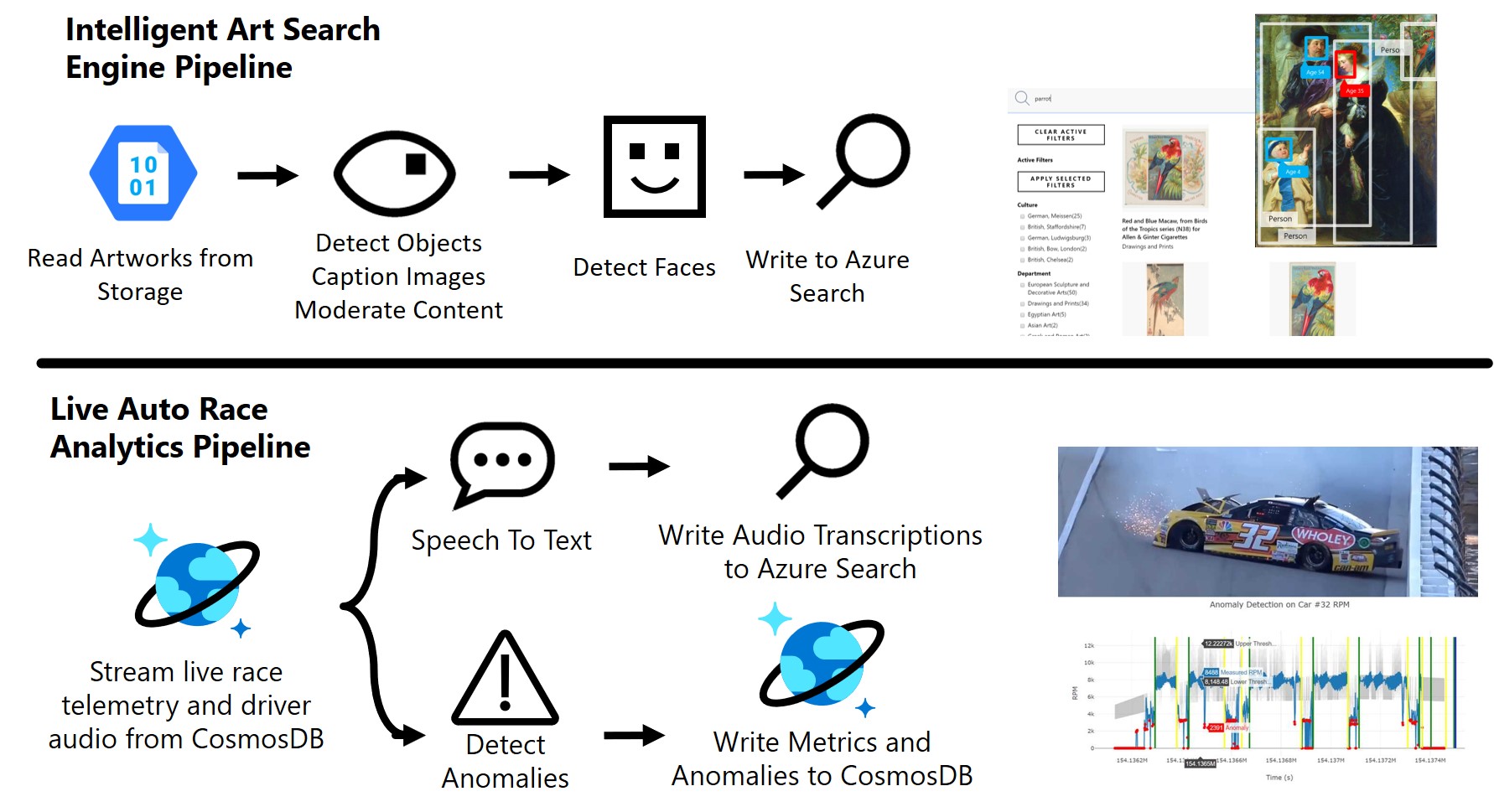}
}
\caption{Top Left: SparkML pipeline of Cognitive Services used to create an intelligent art search engine (pictured top right). Bottom Left: SparkML pipeline of cognitive services for a live race analytics system (pictured bottom right).}
\label{fig:applications}
\end{center}
\end{figure*}

\subsection{Accelerated Networking with Containerized Deployments}

\begin{figure}[ht]
\vskip 0.1in
\begin{center}
\centerline{
\includegraphics[width=\linewidth]{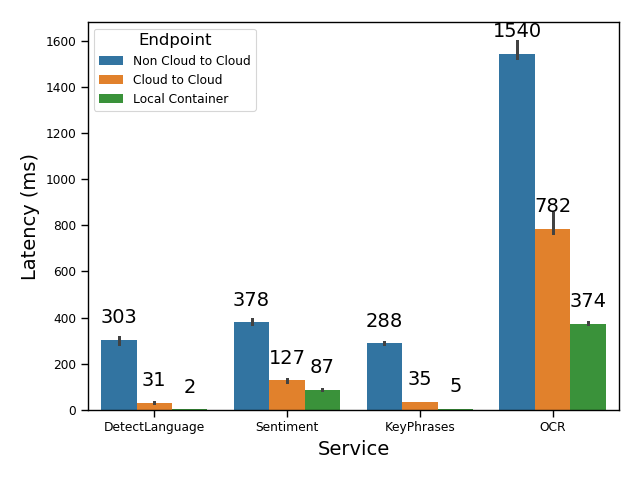}
}
\caption{Request latency as a function of algorithm and deployment type. Local networking with containerized services can provide significantly less latency that inter-machine networking. We also explore the effect of co-locating the client and server in the same cloud region which we refer to as the ``Cloud to Cloud'' deployment. Both outperform the ``Non-Cloud to Cloud'' scenario where the client is outside of the cloud's accelerated network.}
\label{fig:containerized-latency}
\end{center}
\end{figure}

In section \ref{sec:containers} we introduce containerized variants of our intelligent services to enable precision control over network topology and scaling. Using this technology, we provide Helm charts that allow for quick deployment of Spark clusters with locally embedded intelligent services. These local services avoid machine to machine communication latencies and have a dramatic impact on system performance. More specifically, these containerized services can improve latency by orders of magnitude when compared to communicating with the cloud. Using containerized local services allows for Spark to be used as a \textit{micro}-service orchestrator as spark workers and services can be collocated on the same machine. In Figure \ref{fig:containerized-latency} we compare median service latencies across four services and their corresponding containerized variants. In particular, we examine the performance differences between local networking (Local Container), accelerated intra-region cloud networking (Cloud to Cloud), and communication from outside the cloud's accelerated network (Non-Cloud to Cloud). We find that using accelerated networking strategies such as co-locating client and service in the same region or on the same machine has a large effect on the latency and responsiveness of the service. In this analysis, we find that containerized service latency is not noticeably different than the underlying computation time of the intelligent algorithm. This shows that HTTP communication with containerized intelligent services does not have to introduce significant overhead. Furthermore, unlike a tightly coupled function dispatch-based integration, this approach allows flexible architectures, isolation of components, and multi-tenant resource sharing.

\begin{table*}[htbp]
\centering
\caption{Comparison of Sentiment Analysis APIs}
\begin{tabular}{|cc|ccc|ccc|ccc|}
\hline
\multirow{2}{*}{Method}  & \multirow{2}{*}{Dataset} & \multicolumn{3}{c|}{F1}                          & \multicolumn{3}{c|}{Precision}                   & \multicolumn{3}{c|}{Recall}                      \\ \cline{3-11} 
                         &                          & Negative       & Neutral        & Positive       & Negative       & Neutral        & Positive       & Negative       & Neutral        & Positive       \\ \hline
\multirow{10}{*}{Company 2}    & br-pt reviews            & \textbf{0.865} & \textbf{0.873} & \textbf{0.974} & 0.790          & \textbf{0.820} & \textbf{0.758} & \textbf{0.893} & \textbf{0.902} & \textbf{0.884} \\
                         & zh-cn reviews            & -              & -              & -              & -              & -              & -              & -              & -              & -              \\
                         & zh reviews               & -              & -              & -              & -              & -              & -              & -              & -              & -              \\
                         & en reviews               & \textbf{0.851} & \textbf{0.902} & \textbf{0.942} & \textbf{0.866} & \textbf{0.756} & \textbf{0.801} & 0.716          & \textbf{0.894} & 0.858          \\
                         & fr tweets                & \textbf{0.622} & 0.499          & \textbf{0.665} & 0.400          & \textbf{0.770} & 0.725          & \textbf{0.820} & \textbf{0.596} & \textbf{0.589} \\
                         & it reviews               & \textbf{0.826} & \textbf{0.860} & \textbf{0.953} & 0.784          & \textbf{0.776} & \textbf{0.773} & 0.778          & \textbf{0.841} & 0.779          \\
                         & ja reviews               & -              & -              & -              & -              & -              & -              & -              & -              & -              \\
                         & ko reviews               & \textbf{0.739} & 0.741          & 0.781          & 0.705          & \textbf{0.680} & 0.684          & \textbf{0.676} & 0.795          & 0.779          \\
                         & semeval2013 en tweets    & \textbf{0.755} & \textbf{0.729} & \textbf{0.830} & 0.651          & \textbf{0.751} & \textbf{0.733} & \textbf{0.770} & \textbf{0.786} & \textbf{0.775} \\
                         & es tweets                & \textbf{0.663} & \textbf{0.636} & \textbf{0.656} & \textbf{0.618} & \textbf{0.655} & \textbf{0.634} & \textbf{0.678} & \textbf{0.696} & \textbf{0.757} \\ \hline
\multirow{10}{*}{Company 1} & br-pt reviews            & 0.503          & 0.431          & 0.718          & 0.308          & 0.426          & 0.353          & 0.535          & 0.653          & 0.608          \\
                         & zh-cn reviews            & 0.534          & 0.622          & \textbf{0.874} & 0.483          & 0.286          & 0.335          & 0.249          & 0.696          & 0.569          \\
                         & zh reviews               & 0.530          & 0.665          & \textbf{0.899} & 0.528          & 0.225          & 0.240          & 0.212          & 0.699          & 0.583          \\
                         & en reviews               & 0.715          & 0.809          & 0.922          & 0.720          & 0.506          & 0.468          & 0.550          & 0.832          & 0.787          \\
                         & fr tweets                & 0.524          & 0.469          & 0.440          & 0.502          & 0.611          & 0.690          & 0.548          & 0.493          & 0.417          \\
                         & it reviews               & 0.630          & 0.767          & 0.826          & 0.716          & 0.405          & 0.472          & 0.356          & 0.718          & 0.629          \\
                         & ja reviews               & 0.582          & 0.713          & 0.860          & 0.609          & 0.290          & 0.322          & 0.263          & 0.743          & 0.645          \\
                         & ko reviews               & 0.560          & 0.585          & \textbf{0.785} & 0.466          & 0.425          & 0.455          & 0.399          & 0.671          & 0.555          \\
                         & semeval2013 en tweets    & 0.552          & 0.446          & 0.620          & 0.348          & 0.612          & 0.531          & 0.723          & 0.599          & 0.679          \\
                         & es tweets                & 0.553          & 0.506          & 0.578          & 0.450          & 0.542          & 0.506          & 0.584          & 0.612          & 0.617          \\ \hline
\multirow{10}{*}{Ours}   & br-pt reviews            & 0.821          & 0.831          & 0.874          & \textbf{0.792} & 0.759          & 0.715          & 0.810          & 0.872          & 0.876          \\
                         & zh-cn reviews            & \textbf{0.813} & \textbf{0.837} & 0.853          & \textbf{0.821} & \textbf{0.751} & \textbf{0.726} & \textbf{0.778} & \textbf{0.851} & \textbf{0.856} \\
                         & zh reviews               & \textbf{0.686} & \textbf{0.756} & 0.764          & \textbf{0.750} & \textbf{0.554} & \textbf{0.545} & \textbf{0.563} & \textbf{0.749} & \textbf{0.755} \\
                         & en reviews               & 0.773          & 0.832          & 0.866          & 0.800          & 0.647          & 0.579          & \textbf{0.734} & 0.840          & \textbf{0.866} \\
                         & fr tweets                & 0.592          & \textbf{0.543} & 0.481          & \textbf{0.624} & 0.692          & \textbf{0.756} & 0.639          & 0.542          & 0.522          \\
                         & it reviews               & 0.800          & 0.845          & 0.894          & \textbf{0.800} & 0.737          & 0.687          & \textbf{0.795} & 0.820          & \textbf{0.821} \\
                         & ja reviews               & \textbf{0.756} & \textbf{0.802} & \textbf{0.890} & \textbf{0.729} & \textbf{0.676} & \textbf{0.615} & \textbf{0.751} & \textbf{0.789} & \textbf{0.771} \\
                         & ko reviews               & 0.738          & \textbf{0.743} & 0.710          & \textbf{0.780} & 0.673          & \textbf{0.703} & 0.645          & \textbf{0.799} & \textbf{0.818} \\
                         & semeval2013 en tweets    & 0.654          & 0.591          & 0.518          & \textbf{0.687} & 0.688          & 0.716          & 0.662          & 0.683          & 0.744          \\
                         & es tweets                & 0.629          & 0.607          & 0.600          & 0.613          & 0.615          & 0.611          & 0.619          & 0.666          & 0.689          \\ \hline
\end{tabular}
\label{tab:sentiment}
\vspace{-.1in}
\end{table*}

\subsection{Semantic Quality of Intelligent Services}
 In earlier sections we have shown how our architecture allows users to embed cloud intelligence into databases. To many developers it's important to know that these approaches are not just performant, but also semantically relevant to a broad class of datasets and tasks. We demonstrate the semantic quality of a variety of cloud services using standard metrics of Precision, Recall, and F1, for each class \cite{hastie2009elements}. To our knowledge, this is the first comparison of intelligent cloud service quality across a variety of major service providers, which we have kept anonymous as ``Company 1'' and ``Company 2''. In this work, we provide benchmarking numbers for sentiment analysis, face detection, content moderation, anomaly detection and named entity recognition. When possible, we aimed to use evaluation datasets that were large, public, diverse, and commonplace. However, for several tasks we could not find suitable datasets and opted to leverage a third party to acquire a relevant datasets. 
 
 To compare across multiple service providers, we leverage public documentation to call each provider's service, then convert each response to a canonical format for the task labels. We then score the formatted responses using the dataset's labels. We run our semantic quality analysis on the publicly available services during the week of April 20th, 2020 between the 20th and 24th. Cloud providers regularly update services and these results represent a recent snapshot in time.

In Table \ref{tab:sentiment}, we compare sentiment analysis approaches across Microsoft Azure’s Text Analytics v3 API \cite{textanalytics} and comparable services from Company 1 and Company 2. We evaluate these approaches on the SemEval 2013 Task 2 \cite{Nakov_semeval-2013task} dataset, which contains English Twitter data. For a more robust multi-lingual analysis, a third-party company assembled and labeled datasets of reviews and tweets across varying languages. Each dataset labels text with “Positive”, “Negative”, or “Neutral” sentiment.  For each of the APIs, we refer to publicly available documentation from each company to map the outputs into these labels. Microsoft's API returned the direct class to evaluate against.  Our (Microsoft's) API outperforms competitors in Korean, Chinese Traditional, Chinese simplified, and Japanese and Company 2's performance is superior in Brazilian Portuguese, English, French, and Spanish.

In Table \ref{tab:contentmod} we analyze content moderation services. Because of the sensitivity of the task and prevalence of adversaries looking to thwart modern content moderation systems, we use two private datasets with explicit images. Datasets of images were collected from third party organizations and contain three categories: images with nudity (“Adult”), partial nudity (“Racy”), or without explicit content (“Safe”). We used Microsoft’s Content Moderation API \cite{contentmod} and analogous APIs from Company 1 and 2, and normalize the outputs from each APIs to match the categories. Microsoft returned explicit true and false labels for each class. Company 1's approach was superior at detecting safe images with high precision and Racy images with high recall, which many users might find desirable. Microsoft makes a similar, though less pronounced, precision recall trade-off to ensure that the approach does not miss explicit content. Company 2 achieves the best F1 scores with a balanced API but can be more susceptible to true negatives.

In Table \ref{tab:anomdetect} we explore the quality of anomaly detection methods. We leverage a public dataset from Yahoo\cite{yahoo}, and an additional private third-party dataset. Each dataset classifies numeric time series data as anomalous or non-anomalous. We explore the Microsoft Anomaly Detection\cite{anomdetect} service, the Luminol open source library\cite{luminol} (LinkedIn) and the AnomalyDectection\cite{twitter} open source library (Twitter), both OSS libraries. Microsoft has a well-balanced API with a high F1 score and a high precision at detecting anomalies. The Luminol library achieves strong performance in non-anomalous precision and anomalous recall which leads to less missed anomalies.

In Table \ref{tab:facedetect} we explore Face Detection on the open FDDB \cite{fddbTech} dataset. We compare Microsoft’s Face API \cite{faceapi} with face detection APIs from Company 1 and Company 2. We compare each service’s canonicalized bounding boxes with the golden bounding boxes from the FDDB dataset. Because each service has their own definitions of what constitutes a face, such as including hair and other subtleties, we focus this analysis on whether the face is detected as opposed to the bounding box quality. We consider a prediction and a label to ``match'' if they meet an Intersection Over Union (IOU) threshold of $> 0.18$. compared to the golden label. Faces were split into three sets. The first contains all images, the second considers more difficult faces that are \textbf{B}elow a \textbf{M}inimum \textbf{D}etectable \textbf{F}ace \textbf{S}ize of 36 pixels (BMDFS), and the third \textbf{E}xcludes the faces below the MDFS (\textbf{E}MDFS). Microsoft shows a strong performance on larger faces but shows favoring a strong precision on the smaller faces. Company 2 shows a strong ability to detect small faces at the expense of its precision. Company 1 performs the best on the full dataset with good performance for both small and large faces.

In Table \ref{tab:nertable}, we explore textual Named Entity Recognition systems on datasets acquired by third party organizations. We focus on locations, organizations, and people as entities. We use the NEREval library \cite{nereval} to calculate F1, Precision, and Recall. We compared Company 1's named entity recognition API against Microsoft Azure's Text Analytics v3. Microsoft shows competitive precision and excels within the Chinese language. Company 1 tends to have stronger recall and F1 scores.

In aggregate, these results demonstrate the intelligent services we focus on in this work are semantically competitive with the services provided by other major cloud providers. This underscores that our approach can supply not only fast and scalable intelligence but can also provide this without sacrificing quality. We additionally note that our method to create high throughput intelligent service clients could apply to any web service including those of other cloud providers. We welcome these efforts and have open sourced this as part of Microsoft ML for Apache Spark \cite{mmlspark} \link{https://aka.ms/mmlspark}.

\begin{table*}[htbp]
\centering
\caption{Comparison of Face Detection APIs}
\vspace{-.1in}
\begin{tabular}{|cc|ccc|ccc|}
\hline
\multirow{2}{*}{\textbf{Method}} & \multirow{2}{*}{\textbf{Dataset}} & \textbf{}      & \textbf{Recall} & \textbf{}      & \textbf{}      & \textbf{Precision} & \textbf{}      \\ \cline{3-8} 
                                 &                                   & \textbf{BMDFS}  & \textbf{EMDFS}   & \textbf{Total} & \textbf{BMDFS}  & \textbf{EMDFS}      & \textbf{Total} \\ \hline
Company 2                              & fddb                              & \textbf{0.828} & 0.973           & 0.930          & 0.564          & 0.969              & 0.921          \\ \hline
Company 1                           & fddb                              & 0.667          & 0.955           & \textbf{0.961} & 0.694          & \textbf{0.977}     & \textbf{0.954} \\ \hline
Ours                             & fddb                              & 0.511          & \textbf{0.975}  & 0.936          & \textbf{0.787} & 0.959              & 0.950          \\ \hline
\end{tabular}
\label{tab:facedetect}
\vspace{-.1in}

\end{table*}

\begin{table*}[htbp]
\centering
\caption{Comparison of Image Content Moderation APIs}
\vspace{-.1in}
\begin{tabular}{|cc|ccc|ccc|ccc|}
\hline
\multirow{2}{*}{\textbf{Method}} & \multirow{2}{*}{\textbf{Dataset}} & \multicolumn{3}{c|}{\textbf{F1}}               & \multicolumn{3}{c|}{\textbf{Precision}}        & \multicolumn{3}{c|}{\textbf{Recall}}           \\ \cline{3-11} 
                                 &                                   & \textbf{Adult} & \textbf{Racy} & \textbf{Safe} & \textbf{Adult} & \textbf{Racy} & \textbf{Safe} & \textbf{Adult} & \textbf{Racy} & \textbf{Safe} \\ \hline
\multirow{2}{*}{Company 2}             & dataset-1                      & \textbf{0.66}  & \textbf{0.50} & \textbf{0.90} & \textbf{0.53}  & \textbf{0.35} & 0.98          & 0.87           & 0.83          & \textbf{0.83} \\
                                 & dataset-2                     & 0.87           & \textbf{0.56} & \textbf{0.83} & \textbf{0.86}  & \textbf{0.44} & 0.92          & 0.87           & 0.79          & 0.75          \\ \hline
\multirow{2}{*}{Company 1}          & dataset-1                      & 0.23           & 0.20          & 0.24          & 0.13           & 0.11          & \textbf{1.00} & 0.83           & \textbf{0.89} & 0.14          \\
                                 & dataset-2                     & 0.74           & 0.35          & 0.24          & 0.63           & 0.22          & \textbf{1.00} & 0.89           & \textbf{0.85} & 0.14          \\ \hline
\multirow{2}{*}{Ours}            & dataset-1                      & 0.45           & 0.40          & 0.83          & 0.29           & 0.27          & 0.99          & \textbf{0.92}  & 0.79          & 0.72          \\
                                 & dataset-2                     & \textbf{0.88}  & 0.55          & \textbf{0.83} & 0.83           & \textbf{0.44} & 0.96          & \textbf{0.93}  & 0.75          & 0.73          \\ \hline
\end{tabular}
\label{tab:contentmod}
\vspace{-.1in}

\end{table*}

\begin{table*}[htbp]
\centering
\caption{Comparison of Anomaly Detection APIs}
\vspace{-.1in}
\begin{tabular}{|cc|cc|cc|cc|}
\hline
\multirow{2}{*}{\textbf{Method}} & \multirow{2}{*}{\textbf{Dataset}} & \multicolumn{2}{c|}{\textbf{F1}}      & \multicolumn{2}{c|}{\textbf{Precision}} & \multicolumn{2}{c|}{\textbf{Recall}}  \\ \cline{3-8} 
                                 &                                   & \textbf{Negative} & \textbf{Positive} & \textbf{Negative}  & \textbf{Positive}  & \textbf{Negative} & \textbf{Positive} \\ \hline
\multirow{2}{*}{LinkedIn}        & kpi29                             & 0.992             & 0.559             & \textbf{0.994}     & 0.500              & 0.990             & \textbf{0.634}    \\
                                 & yahoo                             & 0.991             & 0.384             & \textbf{0.999}     & 0.254              & 0.984             & \textbf{0.785}    \\ \hline
\multirow{2}{*}{Twitter}         & kpi29                             & 0.988             & 0.411             & 0.993              & 0.323              & 0.982             & 0.566             \\
                                 & yahoo                             & 0.990             & 0.245             & 0.996              & 0.166              & 0.984             & 0.462             \\ \hline
\multirow{2}{*}{Ours}            & kpi29                             & \textbf{0.995}    & \textbf{0.621}    & 0.992              & \textbf{0.798}     & \textbf{0.998}    & 0.508             \\
                                 & yahoo                             & \textbf{0.997}    & \textbf{0.575}    & 0.998              & \textbf{0.518}     & \textbf{0.996}    & 0.646             \\ \hline
\end{tabular}
\label{tab:anomdetect}
\vspace{-.1in}

\end{table*}

\begin{table*}[htbp]
\centering
\caption{Comparison of Named Entity Recognition APIs}
\vspace{-.1in}
\begin{tabular}{|cc|cccc|cccc|cccc|}
\hline
\multirow{2}{*}{\textbf{Method}} & \multirow{2}{*}{\textbf{Dataset}} & \multicolumn{4}{c|}{\textbf{F1}}                                & \multicolumn{4}{c|}{\textbf{Precision}}                         & \multicolumn{4}{c|}{\textbf{Recall}}                            \\ \cline{3-14} 
                                 &                                   & \textbf{All}  & \textbf{Loc}  & \textbf{Org}  & \textbf{Person} & \textbf{All}  & \textbf{Loc}  & \textbf{Org}  & \textbf{Person} & \textbf{All}  & \textbf{Loc}  & \textbf{Org}  & \textbf{Person} \\ \hline
\multirow{10}{*}{Company 1}         & Chinese                           & 0.57          & 0.61          & 0.38          & 0.68            & 0.67          & 0.69          & 0.52          & 0.76            & 0.50          & 0.54          & 0.30          & 0.61            \\
                                 & English                           & \textbf{0.69} & \textbf{0.72} & \textbf{0.58} & 0.75            & \textbf{0.58} & \textbf{0.61} & \textbf{0.50} & 0.64            & \textbf{0.84} & 0.89          & \textbf{0.71} & \textbf{0.92}   \\
                                 & French                            & \textbf{0.72} & \textbf{0.74} & \textbf{0.58} & \textbf{0.78}   & 0.63          & 0.66          & \textbf{0.47} & 0.72            & \textbf{0.83} & \textbf{0.85} & \textbf{0.75} & \textbf{0.85}   \\
                                 & German                            & \textbf{0.63} & \textbf{0.68} & \textbf{0.50} & \textbf{0.71}   & 0.52          & \textbf{0.56} & \textbf{0.40} & 0.62            & \textbf{0.79} & \textbf{0.85} & \textbf{0.69} & \textbf{0.82}   \\
                                 & Italian                           & 0.66          & 0.65          & \textbf{0.49} & \textbf{0.76}   & 0.62          & 0.58          & 0.50          & 0.73            & \textbf{0.71} & \textbf{0.74} & \textbf{0.49} & \textbf{0.80}   \\
                                 & Japanese                          & \textbf{0.58} & \textbf{0.68} & \textbf{0.49} & \textbf{0.50}   & \textbf{0.57} & \textbf{0.64} & \textbf{0.52} & \textbf{0.51}   & \textbf{0.60} & \textbf{0.74} & \textbf{0.47} & \textbf{0.50}   \\
                                 & Korean                            & \textbf{0.61} & \textbf{0.72} & \textbf{0.40} & \textbf{0.62}   & \textbf{0.60} & \textbf{0.78} & 0.35          & \textbf{0.61}   & \textbf{0.61} & \textbf{0.68} & \textbf{0.45} & \textbf{0.64}   \\
                                 & Portuguese                        & \textbf{0.67} & 0.65          & \textbf{0.60} & \textbf{0.82}   & 0.61          & 0.60          & 0.53          & \textbf{0.77}   & \textbf{0.75} & 0.70          & \textbf{0.69} & \textbf{0.88}   \\
                                 & Russian                           & \textbf{0.66} & 0.68          & \textbf{0.51} & \textbf{0.76}   & \textbf{0.66} & 0.65          & \textbf{0.52} & \textbf{0.77}   & \textbf{0.67} & \textbf{0.71} & \textbf{0.50} & \textbf{0.74}   \\
                                 & Spanish                           & \textbf{0.72} & \textbf{0.73} & \textbf{0.59} & \textbf{0.81}   & \textbf{0.63} & \textbf{0.63} & \textbf{0.49} & \textbf{0.75}   & \textbf{0.84} & \textbf{0.87} & \textbf{0.74} & \textbf{0.88}   \\ \hline
\multirow{10}{*}{Ours}           & Chinese                           & \textbf{0.82} & \textbf{0.80} & \textbf{0.80} & \textbf{0.88}   & \textbf{0.86} & \textbf{0.87} & \textbf{0.81} & \textbf{0.89}   & \textbf{0.79} & \textbf{0.73} & \textbf{0.79} & \textbf{0.87}   \\
                                 & English                           & 0.60          & 0.63          & 0.45          & \textbf{0.78}   & 0.47          & 0.48          & 0.33          & \textbf{0.73}   & 0.82          & \textbf{0.90} & \textbf{0.71} & 0.84            \\
                                 & French                            & 0.65          & 0.70          & 0.36          & 0.73            & \textbf{0.70} & \textbf{0.77} & 0.42          & \textbf{0.74}   & 0.61          & 0.65          & 0.31          & 0.73            \\
                                 & German                            & 0.61          & 0.66          & 0.42          & \textbf{0.71}   & \textbf{0.54} & \textbf{0.56} & \textbf{0.40} & \textbf{0.64}   & 0.70          & 0.81          & 0.46          & 0.80            \\
                                 & Italian                           & \textbf{0.69} & \textbf{0.77} & 0.41          & 0.75            & \textbf{0.77} & \textbf{0.81} & \textbf{0.52} & \textbf{0.85}   & 0.62          & \textbf{0.74} & 0.33          & 0.67            \\
                                 & Japanese                          & 0.52          & 0.64          & 0.37          & 0.45            & 0.55          & 0.62          & 0.43          & 0.49            & 0.50          & 0.66          & 0.32          & 0.41            \\
                                 & Korean                            & 0.51          & 0.65          & 0.34          & 0.37            & 0.54          & 0.64          & \textbf{0.44} & 0.40            & 0.49          & 0.67          & 0.28          & 0.34            \\
                                 & Portuguese                        & 0.64          & \textbf{0.71} & 0.49          & 0.72            & \textbf{0.72} & \textbf{0.68} & \textbf{0.73} & 0.76            & 0.57          & \textbf{0.74} & 0.37          & 0.68            \\
                                 & Russian                           & 0.60          & \textbf{0.74} & 0.43          & 0.57            & 0.60          & \textbf{0.78} & 0.47          & 0.53            & 0.59          & 0.70          & 0.41          & 0.63            \\
                                 & Spanish                           & 0.70          & \textbf{0.73} & 0.53          & 0.79            & 0.62          & 0.65          & 0.45          & 0.74            & 0.79          & 0.85          & 0.64          & 0.84            \\ \hline
\end{tabular}
\label{tab:nertable}
\end{table*}

\section{Applications}

Our contributions aim to simplify integration of intelligent algorithms with big data systems. To demonstrate our approach, we describe two production applications. The first is an intelligent search system for the Metropolitan Museum of Art, and the second is a live race analytics platform for a major NASCAR racing team. 

Our intelligent art search engine starts with half a million images from the Metropolitan Open Access Collection. We keep these images in an Azure Storage account and use Spark's Azure Storage Connector. We pipe these images through several vision services to add intelligent insights to each work of art. More specifically, we detect objects and faces, describe the image to improve text-based searches, and classify racy and adult images for child-sensitive downstream applications. Each transformation is an example of our high-throughput cognitive services for big data. We chain these services together using the SparkML pipeline abstraction as shown in Figure \ref{fig:applications}. These transformations enrich the incoming images with intelligent annotations that enable richer search experiences and analysis of art-historical trends such as gender representation through time. We have deployed a variant of this search experience publicly at \link{https://gen.studio}.

To create a real-time race analytics pipeline we use Spark's structured streaming APIs to read engine telemetry and driver audio streams from CosmosDB. We pass the audio streams to the Speech to Text cognitive service and write the results to an Azure Search index. We use this search index to build a human query-able race log to help pit crews and track managers understand the state of the driver, race, and other crew members. We also pass engine telemetry through the Anomaly Detection service and write detections to an additional table in CosmosDB. This collects interesting events for technicians to inspect further and can also power alerting systems. In Figure \ref{fig:applications} we diagram the pipeline and include a screenshot of the anomaly detection interface we expose to race technicians and pit-crews.

\section{Conclusion}

This work presents new architectures and tools to integrate intelligent HTTP services into big data applications simply and efficiently. We leveraged the Apache Spark ecosystem for its broad connectivity and massive elastic parallelism. In particular, we have contributed an integration between the SparkSQL type system and the HTTP Protocol that allows Spark to function as a micro-service orchestrator. We also add a new asynchronous parallelism API which can improve HTTP throughput by orders of magnitude. We build on these contributions to provide easy-to-use high-throughput bindings for the Azure Cognitive Services using a fluent API that integrates with the SparkML ecosystem. We demonstrate the scalability of this approach and performance of these high throughput clients relative to other approaches. To further improve performance and achieve ultra-low latencies, we contributed containerized versions of the Azure Cognitive Services and helm charts to organize their deployment alongside Apache Spark clusters. Finally, we have shown that these algorithms are of competitive quality for a wide variety of general-purpose intelligence tasks spanning text, vision, face, numeric data. 

\section*{Acknowledgments}
The authors of this work would like to acknowledge the many contributors to the Microsoft ML for Apache Spark project including Sudarshan Raghunathan, Ilya Matiach, Markus Cozowicz, Rohit Agrawal, Nisheet Jain and others. We also thank the Microsoft AI Development Acceleration Program including Casey Hong, Manon Knoertzer, Karthik Rajendran, Alejandro Buendia, and Tayo Amuneke for building the Azure Search Cognitive Service for Big Data and many other contributions. We thank Phani Mutyala for his helpful comments on this manuscript, and Henrik Frystyk Nielsen for his instrumental feedback on the early designs of HTTP on Spark. We also acknowledge the Azure Cognitive Services team for supplying the computational resources to explore these challenges and evaluate the proposed solutions. Furthermore we thank the CSAIL Alliances Systems that Learn program for helping to fund this work. 

\FloatBarrier

\bibliographystyle{IEEEtran}
\bibliography{references}

\end{document}